\documentclass[letterpaper, 10 pt, conference]{ieeeconf}  

\IEEEoverridecommandlockouts                              

\usepackage{graphics} 
\usepackage{epsfig} 
\usepackage{mathptmx} 
\usepackage{times} 
\usepackage{amsmath} 
\usepackage{amssymb}  
\usepackage{xcolor}
\usepackage{multirow}
\usepackage{adjustbox}
\usepackage{booktabs}

\usepackage{caption}
\usepackage{cite}
\usepackage{array}
\usepackage{siunitx}
\usepackage{symbols}
\usepackage{color,soul}
\usepackage{float}
\usepackage{stfloats}
\usepackage{graphicx}
\usepackage{hyperref}
\usepackage{tabularx}
\usepackage{subcaption}

\title{\LARGE \bf
TACTFUL: Tactile-Driven Exploration For Object Localization and Identification in Confined Environments
}

\author{Shivani Kamtikar$^{1,2}$, Chung Hee Kim$^{1,3}$, Camilla Tabasso$^{1}$, Tye Brady$^{1}$, Joshua Migdal$^{1}$, Ta\c{s}k{\i}n Pad{\i}r$^{1}$%
\thanks{*This work was done when Shivani Kamtikar and Chung Hee Kim were interns at Amazon.}%
\thanks{$^{1}$Amazon Fulfillment Technologies \& Robotics, Westborough, MA, USA. \texttt{\{bradytye, ptaskin, jmigdal\}@amazon.com}}%
\thanks{$^{2}$Siebel School of Computing and Data Science, University of Illinois at Urbana-Champaign, Champaign, IL, USA. \texttt{skk7@illinois.edu}}%
\thanks{$^{3}$Robotics Institute, Carnegie Mellon University, Pittsburgh, PA, USA. \texttt{chunghek@andrew.cmu.edu}}%
\thanks{Ta\c{s}k{\i}n Pad{\i}r holds concurrent appointments as a Professor of Electrical and Computer Engineering at Northeastern University and as an Amazon Scholar. This paper describes work performed at Amazon and is not associated with Northeastern University.}%
}

\begin{document}

\maketitle
\thispagestyle{empty}
\pagestyle{empty}


\begin{abstract}

Humans effortlessly locate and identify objects by touch alone, even without vision. In contrast, robotic systems rely heavily on vision and struggle with autonomous tactile exploration and object identification. We present TACTFUL, a vision-free tactile exploration framework that enables a multi-fingered robot to autonomously explore confined workspaces, discover objects through contact, and identify them via tactile reconstruction. Trained entirely on real hardware without simulation, our system learns a single policy that balances global workspace exploration with local surface refinement through a dynamic reward schedule. Our results demonstrate that tactile sensing, when paired with structured learning, can serve as an effective primary modality for object-level reasoning, achieving 77\% success with 0.015 m average reconstruction error and outperforming baseline approaches on real-world objects.

\end{abstract}

\section{INTRODUCTION}

Humans possess a remarkable ability to locate and identify objects in visually occluded scenarios, through a combination of tactile sensing and prior geometric knowledge, effortlessly distinguishing between a set of keys and a water bottle in a cluttered bag. Tactile sensors offer direct, meaningful feedback about contact interactions and local surface geometry~\cite{bergmann2019influence, wei2022open, Li2025}, particularly in occluded or confined settings. Yet, robotic manipulation has traditionally relied heavily on vision, with tactile sensing often used as a secondary or corrective modality~\cite{hansen2022visuotactile, huang20243d, bauza2024simple, li2025maniptrans}.  While prior work has explored tactile-based object mapping and identification~\cite{bauza2019tactile, li2020review, shahidzadeh2024actexplore, mao2024dexskills}, autonomous tactile exploration for object-level reasoning remains underexplored in real-world settings. Moreover, most prior tactile systems have primarily been evaluated in simulation using simplified or binary contact signals, with comparatively limited work demonstrating real-world performance using high-fidelity tactile data~\cite{xu2022tandem3d, pai2023tactofind}. Consider a robot exploring a confined workspace containing multiple objects. When visual perception is available, object localization and identification are relatively straightforward, as camera-based systems can provide dense scene observations. In vision-denied scenarios, robots must rely on local tactile feedback to explore workspaces, localize objects, and reconstruct geometry. This requires simultaneously addressing workspace exploration for object discovery and contact-driven surface reconstruction for identification.

\begin{figure}[htbp]
    \centering
    \includegraphics[scale=0.46]{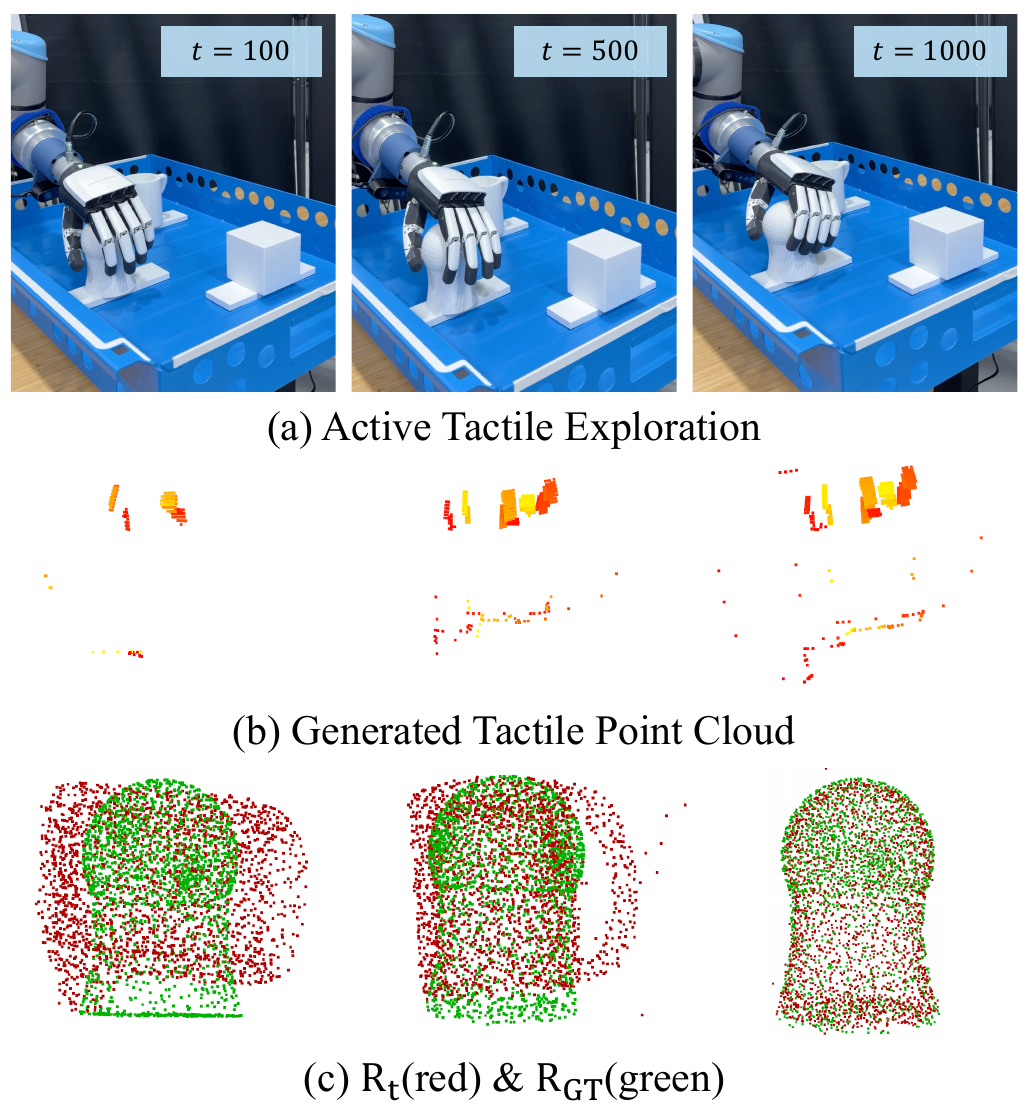}
    \caption{(a) Active tactile exploration by the agent at various timesteps (no vision), (b) tactile point cloud accumulated at timestep $t$, and (c) shape-completed reconstruction, $R_t$ (red), overlaid on the ground truth geometry, $R_{GT}$ (green)}
    \label{fig:task}
    \vspace{-13pt}
\end{figure}

We present \ourmethod, a real-world, vision-free tactile exploration and identification framework for settings where objects are drawn from a known object library but must be located without vision, such as warehouse bin-picking, medical instrument retrieval, or in manufacturing industries. Our framework makes the following contributions: 1) we develop a learned tactile exploration policy that enables a multi-fingered robot to autonomously \textit{explore} and \textit{localize} objects within confined workspaces, 2) we introduce a dynamic reward schedule that balances global exploration with local contact refinement, leveraging object-level geometric priors to guide 3D reconstruction, and finally, 3) we demonstrate reliable object \textit{identification} based solely on reconstructed tactile geometry.



\section{Related Work}
\begin{figure*}[htp!]
\centering
\smallskip
  \includegraphics[width=1\textwidth]{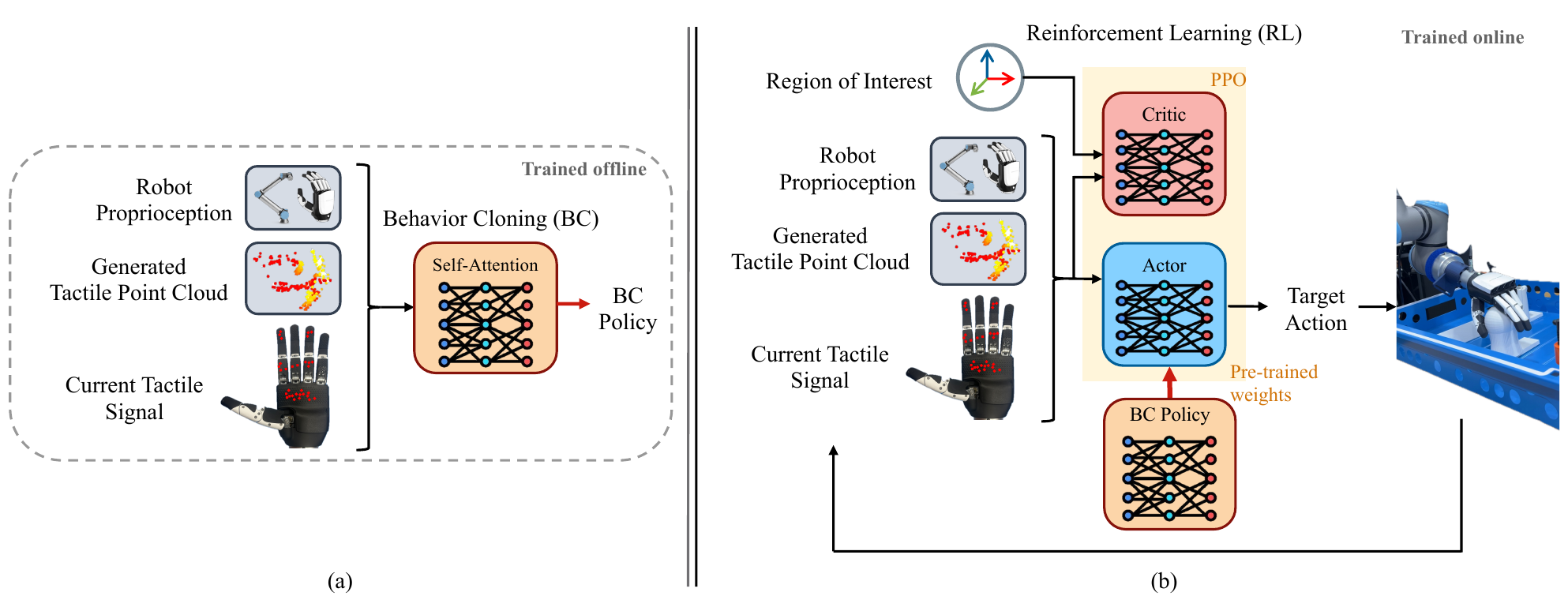}
  \caption{\textbf{Pipeline of \ourmethod:} (a) Offline behavior cloning (BC) model trained using proprioception, tactile signals, and tactile point cloud; used to initialize the actor in the PPO-based reinforcement learning (RL) policy. (b) Online PPO training with the same inputs and an additional region of interest (ROI) for the critic, resulting in an exploration policy for object-guided reconstruction in confined spaces.}
  \label{fig:pipeline}
  \vspace{-.5cm}
\end{figure*}
\noindent\textbf{Tactile-based localization.}
Tactile sensing has been explored as a means for object localization when visual information is unreliable or unavailable. Early work demonstrated tactile SLAM using whisker-inspired sensors~\cite{fox2012tactile}, while subsequent approaches employed simple tactile probes for active localization~\cite{dragiev2013uncertainty, schneiderapple, yi2016active}. However, these methods typically rely on low-dimensional contact signals and lack the rich spatial information afforded by multi-fingered tactile hands. Kissoum~\textit{et al.}~\cite{kissoum2021simultaneous} addressed simultaneous localization and reconstruction using particle filters, but their approach is limited to 2D simulated environments with simplified contact dynamics. In contrast, our work leverages high-resolution, real-world tactile sensing to jointly localize and reconstruct objects through learned exploration policies.

\noindent\textbf{Object reconstruction using tactile data.}
Reconstructing object geometry from touch has gained increasing attention in vision-denied settings, with some works exploring efficient shape exploration using rigid tactile arrays~\cite{fleer2020learning} and learning-based methods that fuse contact positions and normals to build 3D object representations~\cite{xu2022tandem3d}. Other approaches focus on reconstructing visually challenging objects, including transparent objects~\cite{murali2023touch}. More recent systems, such as TactoFind~\cite{pai2023tactofind}, demonstrate tactile-only object retrieval in clutter, but depend on hand-crafted exploration heuristics rather than learned policies. However, the majority of tactile reconstruction methods assume that the object has already been localized and focus exclusively on surface exploration. Our work addresses a more general and realistic setting in which the robot must first explore an unknown workspace to discover objects and then perform detailed tactile interaction to reconstruct their geometry, requiring policies that tightly couple global exploration with local shape refinement.

\noindent\textbf{Deep reinforcement learning for tactile perception.}
Deep reinforcement learning has been widely applied to tactile-driven perception and exploration tasks. Prior work has highlighted the importance of memory for sequential tactile decision-making, using recurrent architectures such as LSTM-augmented A3C~\cite{mirowski2016learning, ramani1904short}. More recent approaches leverage attention mechanisms to generalize tactile perception across manipulation primitives~\cite{schneiderapple}. Similarly, Lee ~\textit{et al.}~\cite{lee2024robot} apply reinforcement learning for occluded object retrieval, but focus on reactive behaviors rather than systematic object reconstruction. In the context of exploration, intrinsic motivation strategies such as D-optimality~\cite{placed2020deep} and contact-based exploration bonuses~\cite{shahidzadeh2024actexplore} have been proposed to encourage informative interactions. However, these methods typically operate under the assumption that object locations are known and focus on localized surface exploration. In contrast, our approach learns a single tactile-only policy that explores an entirely unknown workspace, localizes objects through contact, and incrementally reconstructs their geometry in the real world using a structured reward design that balances search, interaction, and shape understanding.

\section{Tactile Robotic Setup}
Our system, shown in Fig.~\ref{fig:task}, consists of a five-fingered Inspire RH56DFTP Series Dexterous hand mounted on a UR10e robotic arm. The robotic hand is equipped with resistive 1062 high-resolution taxels on the fingertips, fingerpads, and the palm area, measuring normal contact forces. The workspace consists of a bin with known bounds having 3D-printed objects placed in arbitrary locations. We do not use any cameras.

\section{Method}
We address vision-free object localization and identification in confined workspaces containing multiple objects. Given a bin of objects drawn from a known library, the robot is assigned a target object and must locate it using only tactile feedback. During exploration, it may encounter distractor objects and must distinguish the target through accurate tactile reconstruction. This requires jointly exploring the workspace to discover objects via contact and reconstructing surface geometry with sufficient fidelity for reliable identification. \ourmethod~learns a single reinforcement learning (RL) policy that enables a multi-fingered robot to autonomously explore and identify objects using tactile sensing alone. The policy is trained with a dynamic reward schedule that balances global workspace exploration with local, contact-rich surface refinement. At the start of each episode, object locations are unknown, requiring active exploration through continuous end-effector and finger motions. As the robot moves, it records end-effector poses and accumulates tactile contacts, forming a sparse tactile point cloud of the contacted surface. Once sufficient data is collected, the sparse point cloud is passed to a learned shape completion model to produce a dense reconstruction, which is then matched against a known object library for identification.

\subsection{Behavior Cloning Initialization}
Learning tactile exploration policies directly through RL is particularly challenging in real-world settings, where random exploration is inefficient and can lead to unsafe contacts~\cite{8794102, 10.5555/3398761.3398819}. Furthermore, the lack of high-fidelity tactile simulations~\cite{xu2023efficient} prevents reliable pre-training in simulation. To overcome these limitations, we initialize the RL policy using Behavior Cloning (BC)~\cite{fang2019survey, 9289148}, allowing it to learn structured exploration behaviors from 5311 tactile-only teleoperated demonstrations, consistent with our vision-free policy. During BC training, the policy is trained to regress the expert action, represented as a concatenation of delta end-effector pose and delta hand configuration, given the current observation state. The BC policy training pipeline is shown in Fig. \ref{fig:pipeline}(a). The loss function ($L_{\text{BC}}$) is formulated as a mean squared error (MSE) between predicted and expert action deltas:

\begin{equation}
L_{\text{BC}} = \frac{1}{N}\sum_{i=1}^{N}\lVert a_i^{\text{pred}} - a_i^{\text{expert}}\rVert_2^2,
\end{equation}

where \(a_i^{\text{pred}}\) and \(a_i^{\text{expert}}\) denote the predicted and expert actions (detailed in Section~\ref{subsection:RL}) at sample \(i\), respectively, and \(N\) is the number of demonstration samples. The resulting BC policy captures basic contact strategies and local exploration patterns demonstrated by the expert. This pretrained policy is then used to initialize the RL policy (Section \ref{subsection:RL}). 

\subsection{POMDP Formulation}
\label{subsection:RL}
We formulate the tactile exploration task as a Partially Observable Markov Decision Process (POMDP) defined by the tuple $M = (S, A, T, \mathcal{O}, \Omega, R, \gamma)$, where $S$ is the underlying state space, $A$ is the action space, $T: S \times A \rightarrow S$ is the state transition function, $\mathcal{O}$ is the observation space, $\Omega: S \rightarrow \mathcal{O}$ is the observation function mapping states to observations, $R: S \times A \rightarrow R$ is the reward function, and $\gamma \in [0,1]$ is the discount factor. The true state includes the complete 3D geometry and pose of all objects in the workspace, information that is not directly observable to the agent. Instead, the agent must infer object locations and shapes through accumulated tactile feedback over time. At each timestep $t$, the agent receives an observation $o_t = \Omega(s_t) \in \mathcal{O}$ consisting of proprioceptive information (end-effector pose, joint angles, hand configuration) and high-resolution tactile sensor readings from the 1062 taxels distributed across the robot hand. These observations provide only local, contact-based information about the environment, making the process partially observable. The goal is to learn a policy $\pi: \mathcal{O}^k \rightarrow A$ that maps a history of $k$ recent observations to actions that maximize the expected cumulative discounted reward:

\begin{equation}
\pi^* = \arg\max_{\pi} \mathbb{E}_{\tau \sim \pi} \left[ \sum_{t=0}^{T} \gamma^t r_t \right]
\end{equation}

\noindent where $\tau = (o_0, a_0, r_0, o_1, a_1, r_1, \ldots)$ is a trajectory sampled under policy $\pi$, and $T$ is the episode horizon. We solve this POMDP using Proximal Policy Optimization (PPO)~\cite{schulman2017proximal}, an on-policy actor-critic reinforcement learning algorithm.

\noindent\textbf{Observation Space ($\mathcal{O}$):} At timestep $t$, the agent observes $o_t \in \mathcal{O}$ consisting of:
\begin{itemize}
    \item \textbf{Tactile readings} $f_t \in \mathbb{R}^{1062}$: Force measurements from each taxel, indicating contact locations and intensities across the hand surface.
    \item \textbf{Tactile Point Cloud $c_t \in \mathbb{R}^{1088}$}\noindent: Learned embedding of the accumulated tactile point cloud history obtained through Sonata~\cite{wu2025sonata}, a self-supervised transformer-based point cloud encoder. This embedding encodes both spatial coverage and geometric structure of the explored surface across the entire episode history, providing the agent with a compressed representation of its exploration progress.
    \item \textbf{End-effector pose} $p_t \in \mathbb{R}^9$: Normalized position and sine-cosine encoded orientations.
    \item \textbf{Joint angles} $j_t \in \mathbb{R}^6$: Normalized arm joint configurations from forward kinematics.
    \item \textbf{Hand configuration} $h_t \in \mathbb{R}^6$: Normalized finger joint angles from forward kinematics.
\end{itemize}
At each timestep $t$, a single observation $o_t \in \mathbb{R}^{2171}$ is constructed as:

\begin{equation}
o_t = [p_t, j_t, h_t, f_t, c_t] \in \mathbb{R}^{2171}
\label{eq:state}
\end{equation}

\noindent\textbf{State Representation ($s_t$):} 
Since individual observations provide insufficient context for long-horizon tactile exploration (e.g., determining movement direction along a surface requires temporal information), the state is represented as a sequence of $k = 10$ recent observations:

\begin{equation}
    s_t = o_t, o_{t-1},..., o_{t-(k-1)}
\end{equation}

This temporal sequence is processed by a transformer architecture (described in \ref{section:training}), which naturally handles the sequential dependencies through its self-attention mechanism, eliminating the need for explicit recurrent connections.

\noindent\textbf{Action Space ($A$):} The action space is defined as:

\begin{equation}
a_t = [\Delta p_t, \Delta h_t] \in \mathbb{R}^{12}
\end{equation}

\noindent where $\Delta p_t \in [-1, 1]^6$ represents incremental changes to the end-effector pose, and $\Delta h_t \in [-1, 1]^6$ represents changes to the finger joint angles. Actions are normalized using z-score standardization to stabilize training.

\noindent\textbf{Partial Observability:} A challenge in our setting is that the agent never observes the true object locations, geometries, or the completeness of its exploration. Without vision, the agent must rely entirely on the history of tactile contacts and proprioceptive feedback to build an implicit understanding of the workspace structure, necessitating memory mechanisms (the point cloud embedding $c_t$ and observation history) to enable effective exploration and reconstruction.

\begin{figure}[h]
    \centering
    \includegraphics[height=0.23\textheight, keepaspectratio]{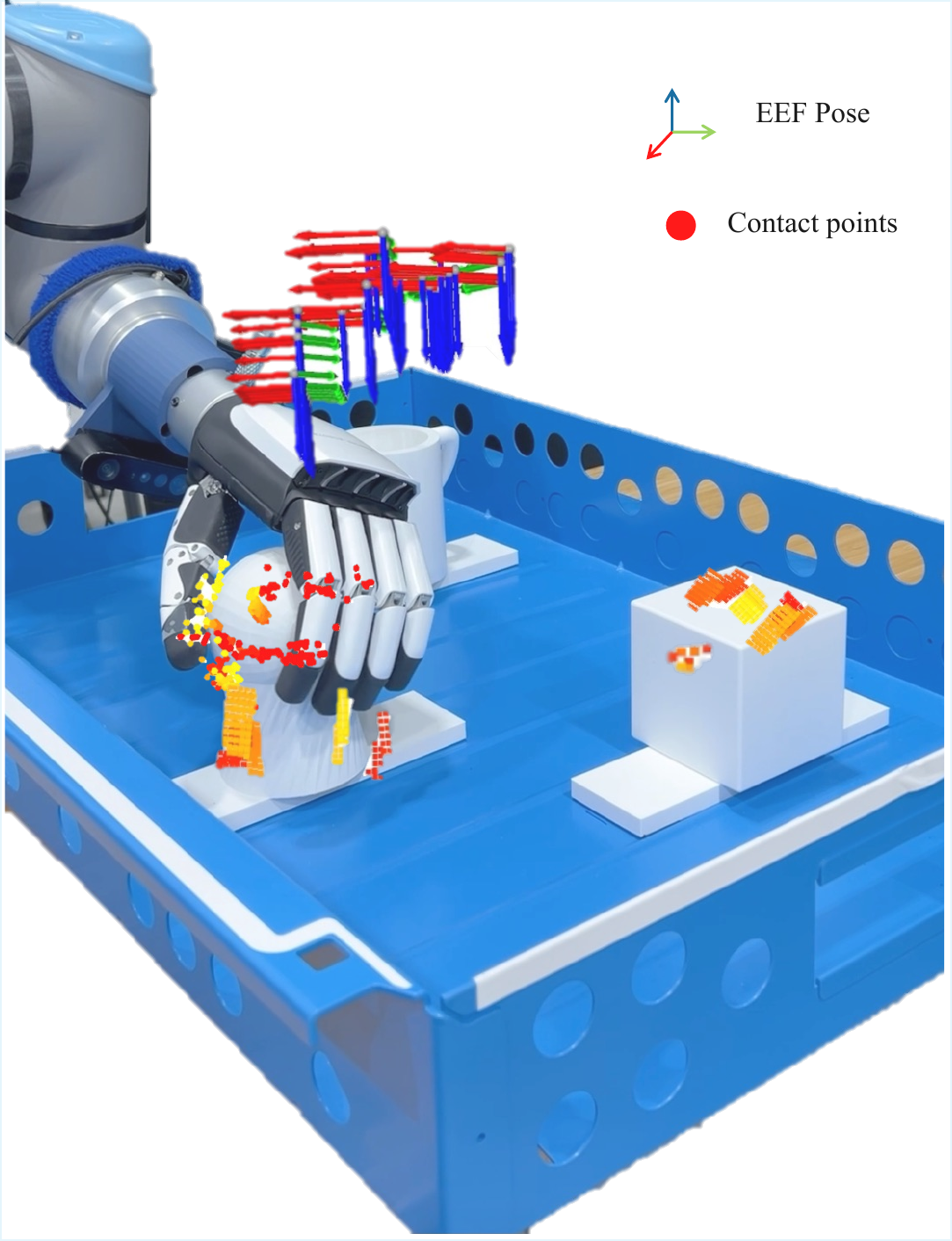}
    \caption{Example map and the tactile point cloud generated during exploration, with the agent tracking end-effector (EEF) poses and contact regions.}
    \label{fig:map}
    \vspace{-12pt}
\end{figure}

\subsection{Dynamic Reward Design}
We design a composite reward function to guide the agent toward effective tactile exploration, shape reconstruction, and diverse surface coverage. The different reward components are explained below.

\noindent\textbf{Contact Reward ($r_{\text{contact}}$):} Promotes informative surface exploitation by encouraging the agent to refine contact around already discovered objects through novel, spatially distinct tactile interactions. \noindent For each activated taxel, we compute its 3D position in world coordinates by combining the taxel's local position on the hand surface with the current end-effector pose $p_t$ and hand configuration $h_t$. We also maintain a history of all contact points accumulated throughout each episode. By referencing this history, actions that create new, previously unseen contact points are incentivized rather than redundant or repetitive touches. A new contact point is considered novel if its Euclidean distance to all previously accumulated points exceeds a minimum distance threshold of 5 mm. It is computed as the number of new, previously unseen points in the reconstruction during time step \( t \), normalized by the maximum possible number of points:

\begin{equation}
    \label{eq:contact_reward}
    r_{\text{contact}} = \frac{\text{diff}_t}{N_{\max}}
\end{equation}

\noindent where \( \text{diff}_t \) denotes the number of new points, and \( N_{\max} \) is the normalization constant.
    
\noindent\textbf{Exploration Reward ($r_{\text{explore}}$):} Encourages the agent to explore new, previously unvisited regions in the workspace. It incentivizes the agent to explore more areas in the confined tray. We track the agent's trajectory by keeping a memory of all visited end effector (EEF) poses. We discretize the workspace into a $1cm^3$ voxel grid and mark a voxel as explored when the end-effector center enters it. By incorporating this history, actions that lead to the agent exploring previously unvisited areas in the tray are rewarded prominently. This also helps the agent to prevent getting stuck in local minima by exploring the same region repeatedly. Adding this reward ensures that the agent has a broader understanding of the workspace and, in turn, helps in convergence. $r_{\text{explore}}$ is defined as:
\begin{equation}
    r_{\text{explore}} = \frac{V_t^{\text{new}}}{V_{\max}},
\end{equation}
    where \( V_t^{\text{new}} \) is the number of new voxels explored at time \( t \), and \( V_{\max} \) is the maximum number of voxels in the grid.

\noindent\textbf{Reconstruction Reward ($r_{\text{recon}}$):} Encourages the agent to produce accurate and complete object reconstructions as it explores by penalizing imprecise or incomplete surface representations. Specifically, we use the Chamfer distance, which measures the average closest-point discrepancy between two point sets, to quantify reconstruction quality. A lower Chamfer distance indicates better alignment and higher fidelity to the true object shape. By incorporating this metric, the agent is incentivized to exploit gathered tactile data to refine local geometry and close coverage gaps, leading to more precise object understanding. This reward thus complements $r_{\text{contact}}$ and $r_{\text{explore}}$: while the $r_{\text{contact}}$ promotes local, high-resolution coverage and $r_{\text{explore}}$ encourages global scene coverage, $r_{\text{recon}}$ ensures that the overall shape representation remains faithful to the true object geometry. At each timestep $t$, the accumulated sparse tactile point cloud is passed through a pretrained shape completion network (detailed in \ref{shape completion}) to generate a dense reconstructed point cloud $R_t$. This reconstruction is compared against the ground-truth point cloud $R_{GT}$ of the target object using normalized Chamfer distance:

\begin{equation}
    r_{\text{recon}} = \frac{\text{Chamfer}(R_t, R_{\text{GT}})}{\text{Chamfer}_{max}},
\end{equation}

\noindent The total reward at each time step \( t \) is given by:

\begin{equation}
r_t = \alpha_t \, r_{\text{contact}} + \beta_t \, r_{\text{explore}} - \lambda_t \, r_{\text{recon}},
\end{equation}

\noindent where \( \alpha_t, \beta_t, \lambda_t \in \mathbb{R}^{+} \) are the time-varying weighting coefficients. 

\noindent To balance global workspace exploration early in an episode with local surface refinement later, we employ a time-varying reward schedule in which the weights of the reward components evolve as a function of the timestep $t$ and episode horizon $T$. Specifically, we use a linear schedule:

\begin{align}
\alpha(t) &= \alpha_{\text{min}} + (\alpha_{\text{max}} - \alpha_{\text{min}}) 
\cdot \frac{t}{T} \\
\beta(t) &= \beta_{\text{max}} - (\beta_{\text{max}} - \beta_{\text{min}}) 
\cdot \frac{t}{T} \\
\lambda(t) &= \lambda_{\text{min}} + (\lambda_{\text{max}} - \lambda_{\text{min}}) 
\cdot \frac{t}{T}
\end{align}

\noindent Here $\beta(t)$ initially prioritizes workspace coverage to facilitate object discovery through incidental contact, while $\alpha(t)$ and $\lambda(t)$ progressively increase to emphasize informative surface interaction and penalize inaccurate reconstructions once contact is established. We set $\alpha_{\text{min}} = 0.1$, $\alpha_{\text{max}} = 0.5$, $\beta_{\text{min}} = 0.1$, $\beta_{\text{max}} = 0.7$, 
$\lambda_{\text{min}} = 0.2$, and $\lambda_{\text{max}} = 0.6$. Although a single policy is used throughout an episode, this evolving reward landscape induces a behavioral transition from broad exploratory motions to deliberate, contact-rich interactions around discovered objects. Importantly, $r_{\text{explore}}$ remains active throughout the episode, ensuring that late object discovery does not prevent successful reconstruction.

\subsection{Training Procedure}
\label{section:training}
The RL policy, shown in Fig. \ref{fig:pipeline}(b), is trained entirely using data collected on the hardware setup with no simulation. Both the actor (policy) and critic (value) networks share the same architecture, consisting of two transformer layers with an embedding dimension of 512 and 4 attention heads. An adaptive average pooling layer is applied to condense the sequence of encoded observations into a global representation, allowing the model to capture long-range temporal dependencies and high-dimensional tactile data into a compact embedding. We train for 10 episodes with a maximum horizon of 2000 steps per episode. We use standard PPO hyperparameters: batch size of 16, initial learning rate of 0.001 with linear decay, clipping $\epsilon = 0.2$, discount factor $\gamma = 0.99$ and $0.016$ KL threshold to ensure stable policy updates. The observation history length is k = 10, and the workspace voxel grid uses $1cm$ resolution for exploration tracking. To stabilize value estimation and guide learning in early stages, we apply \textit{asymmetric observation} between the policy and value networks~\cite{mirowski2016learning, lee2024robot}. The value network is provided with additional privileged information - the approximate 6-DoF pose (within 0.10 m) of the target object (region of interest as shown in Fig. \ref{fig:pipeline}(b)). This is not accessible to the policy network during deployment. Note that when multiple objects are present in the workspace, the critic only observes the target object's approximate pose, not the poses of distractor objects.

\subsection{Shape Completion}
\label{shape completion}
To transform sparse tactile measurements into complete object representations, we employ a point cloud diffusion model \cite{romanelis2024efficient} trained to perform shape completion from partial observations. The model is pretrained on a diverse object set and learns to infer full object geometry from partial, contact-based point clouds. Given tactile measurements, it predicts a dense reconstruction that can be directly compared against the target object. This model effectively serves as a learned prior over our known object set, allowing us to bridge the gap between sparse tactile observations and complete object geometries.

\section{Experiments and Results}

\subsection{Task Formulation}
\begin{figure}[htbp]
    \centering
    \includegraphics[width=0.9\linewidth]{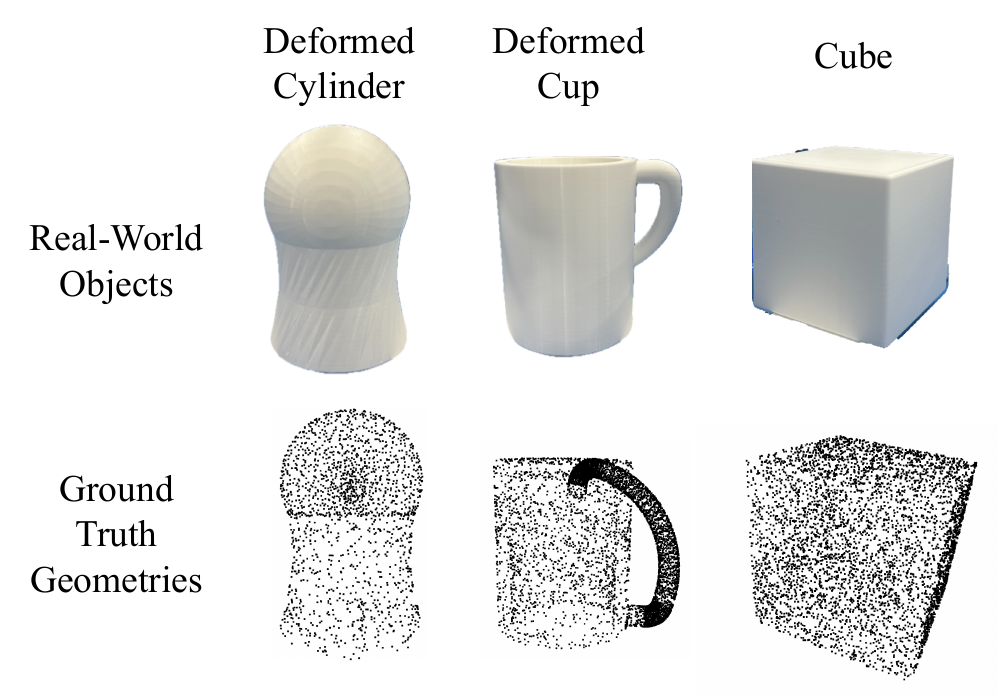}
    \caption{Real-world objects used for the experiments, along with the ground truth point clouds for each of them.}
    \label{fig:objects}
    \vspace{-10pt}
\end{figure}

We evaluate our framework on three geometrically diverse real-world objects 
(Fig.~\ref{fig:objects}), selected based on graspability metrics~\cite{wang2019adversarial}: 
a \textit{cube} (high graspability), a \textit{deformed cylinder}, and a 
\textit{deformed cup} (lower graspability). In each episode, all three objects 
are placed in a bin with known spatial bounds at randomized positions. The robot 
is assigned one of the three objects as its target.

\noindent\textbf{Episode Termination:} An episode terminates under three conditions: 
(1) \textit{Success}: the agent correctly identifies the target object (shape 
completion achieves Chamfer distance below 0.02 m for 10 consecutive timesteps), 
(2) \textit{Failure}: the agent incorrectly identifies a non-target object with 
high confidence, or (3) the maximum horizon T = 2000 steps is reached. Safety 
violations (workspace boundary exceeded or collision detected) also trigger 
termination and are counted as failures.

\noindent\textbf{Metrics:} The tests are evaluated based on 2 different success metrics: 

\noindent\textit{1) Success Rate}: This is the success rate over trials. A trial is a success if the agent identifies and locates the target object. 

\noindent\textit{2) Chamfer-$L_2$}: We measure the Chamfer-$L_2$ distance between the generated (shape-completed) and ground-truth point clouds. Let \( P_{\text{GT}} = \{ \mathbf{p}_i \}_{i=1}^{N} \) be the set of points in the aligned ground truth point cloud, and \( P_{\text{t}} = \{ \mathbf{q}_j \}_{j=1}^{M} \) be the set of points in the reconstructed point cloud. The one-way Chamfer distances are defined as: 
    \begin{equation}
    \text{Chamfer}_{\text{GT} \rightarrow \text{t}} = \frac{1}{N} \sum_{i=1}^{N} \min_{\mathbf{q}_j \in P_{\text{t}}} \| \mathbf{p}_i - \mathbf{q}_j \|_2
    \end{equation}
    
    \begin{equation}
    \text{Chamfer}_{\text{t} \rightarrow \text{GT}} = \frac{1}{M} \sum_{j=1}^{M} \min_{\mathbf{p}_i \in P_{\text{GT}}} \| \mathbf{q}_j - \mathbf{p}_i \|_2
    \end{equation}
    The full symmetric Chamfer distance is the sum of both directions:
    \begin{equation}
    \text{Chamfer}(P_{\text{GT}}, P_{\text{t}}) = \text{Chamfer}_{\text{GT} \rightarrow \text{t}} + \text{Chamfer}_{\text{t} \rightarrow \text{GT}}
    \end{equation}
    
\noindent\textbf{Baselines (All results detailed in Section \ref{section:results} and Table~\ref{tab:main_results})-} We do not compare against visuotactile baselines (e.g.,~\cite{hansen2022visuotactile,bauza2024simple, li2025maniptrans}) 
as they assume visual observability, which directly contradicts our vision-denied setting. Similarly, purely vision-based methods are inapplicable. We instead compare against tactile-only approaches, which operate under identical sensory constraints.

\noindent\textit{1) Heuristic Exploration Policy (Heuristic)}: For the heuristic baseline, we implemented a policy similar to TactoFind~\cite{pai2023tactofind}, with the addition of more hand configurations like full grasp, partial grasp, and wrist movements. The policy first locates the objects within the voxelized bin to map them, then systematically explores them to obtain fine-grained tactile data for identification.  

\noindent\textit{2) Behavior Cloning (BC only)}: We use only our BC policy without the addition of the RL policy and rewards.

\noindent\textit{3) Reinforcement Learning without Behavior Cloning (RL w/o BC)}: We train the RL policy without a BC initialization. 

\noindent\textbf{Effect of Various Reward Functions:} We conduct an ablation study to evaluate the contribution of each reward component in our PPO training framework. We train the PPO model with only 2 reward functions at a time and conduct experiments on all three objects using this model. Results detailed in~\ref{section:results} and Table~\ref{tab:rewards}.

\noindent\textbf{Effect of Shape Completion Model:}
We ablate the shape completion model by removing it and comparing the raw sparse point cloud to the ground truth, evaluating its impact on success rate and Chamfer distance. Results detailed in~\ref{section:results} and Table \ref{tab:shape}.

\begin{figure*}[htp!]
\centering
\smallskip
  \includegraphics[width=0.95\linewidth]{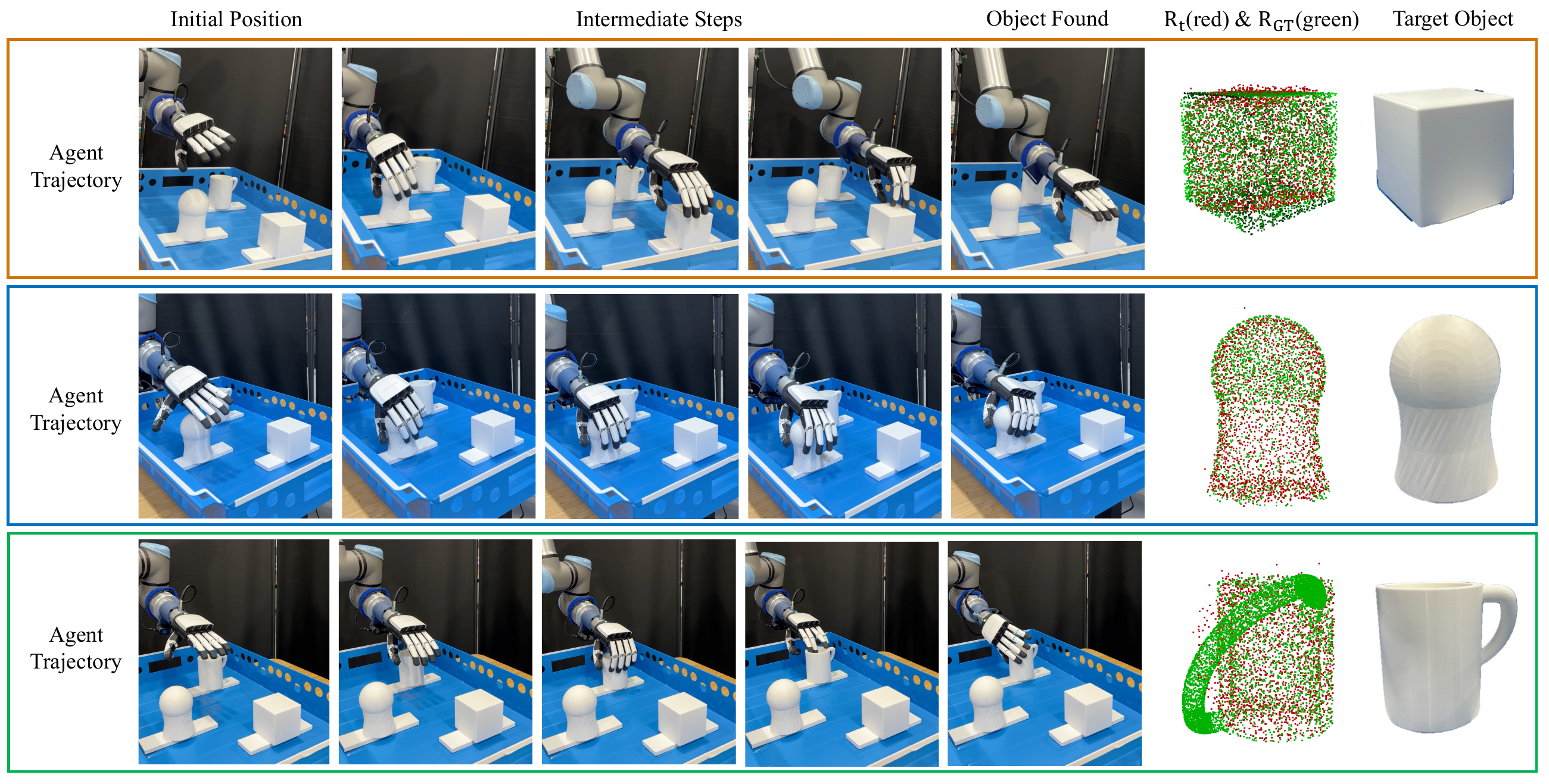}
  \caption{\textbf{Experimental results (one example test run for each of the objects)}: Shows initial view, three intermediate views, and the final view (end of test) of the exploration sequence. The shape-completed reconstruction, $R_t$ (red), overlaid on the ground truth geometry, $R_{GT}$ (green), is also shown. The final column shows the goal object that the robot was tasked to identify. (Also see supplementary video.)}
  \label{fig:exp}
  \vspace{-.3cm}
\end{figure*}

\begin{table*}[htp!]
\centering
\caption{\textbf{Quantitative results on exploration and object identification.} Each row shows results when that object is the designated target, with all three objects present in the workspace. The table shows the average success rate over 12 trials for every object is shown along with the Chamfer distance (m) (Mean $\pm$ STD) for all methods. Additionally, the initial distance from the target (Init. $\Delta$ (m)) is also given.}
\resizebox{\textwidth}{!}{%
    \begin{tabular}{l@{\hskip 4mm} c c c c c c c c c c}
    \toprule
    & & \multicolumn{3}{c}{Deformed Cylinder} & \multicolumn{3}{c}{Cube} & \multicolumn{3}{c}{Deformed Cup} \\
    \cmidrule(lr){3-5} \cmidrule(lr){6-8} \cmidrule(lr){9-11}
    \# & Method & Success $\uparrow$ & Chamfer-$L_2$ $\downarrow$ & Init. $\Delta$ & Success $\uparrow$ & Chamfer-$L_2$ $\downarrow$ & Init. $\Delta$ & Success $\uparrow$ & Chamfer-$L_2$ $\downarrow$ & Init. $\Delta$ \\
    \midrule
    1 & Heuristic & 0.416 & 0.050$\pm$0.023 & 0.230$\pm$0.029 & 0.5 & 0.052$\pm$0.020 & 0.318$\pm$0.004 & 0.416 & 0.130$\pm$0.100 & 0.334$\pm$0.009\\
    2 & BC only & 0.416 & 0.044$\pm$0.015 & 0.246$\pm$0.062 & 0.583 & 0.063$\pm$0.007 & 0.253$\pm$0.063 & 0.583 & 0.120$\pm$0.100 & 0.331$\pm$0.071\\
    3 & RL w/o BC & 0.750 & 0.020$\pm$0.001 & 0.201$\pm$0.041 & 0.750 & 0.023$\pm$0.003 & 0.256$\pm$0.048 & 0.583 & 0.031$\pm$0.015 & 0.321$\pm$0.061\\
    4 & \textbf{Ours} & \textbf{0.833} & \textbf{0.010$\pm$0.004} & \textbf{0.214$\pm$0.051} & \textbf{0.833} & \textbf{0.012$\pm$0.009} & \textbf{0.255$\pm$0.059} & \textbf{0.667} & \textbf{0.021$\pm$0.011} & \textbf{0.322$\pm$0.068}\\
    \bottomrule
    \end{tabular}}
\label{tab:main_results}
\vspace{-.5cm}
\end{table*}

\subsection{Experimental Results}
\label{section:results}
\begin{figure*}[htp!]
\centering
\smallskip
  \includegraphics[width=1\linewidth]{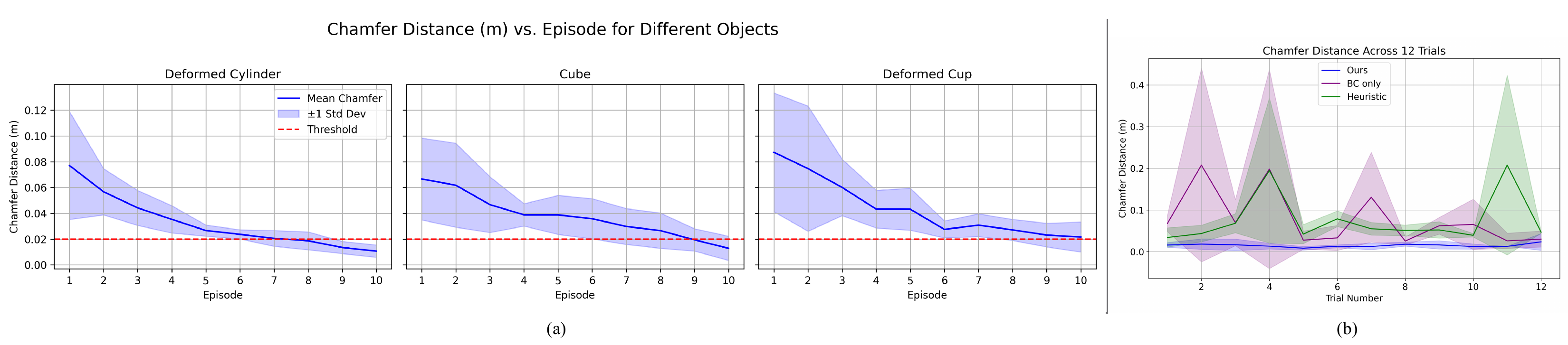}
  \caption{\textbf{(a) Chamfer distance vs Episode of our method across three objects}: Each curve shows the mean Chamfer distance over 10 episodes, with shaded regions representing $\pm$ 1 standard deviation. The decreasing trend shows improved reconstruction accuracy over training. ; \textbf{(b) Chamfer distance comparison for various policies}: Plot shows the mean Chamfer distance over 12 trials for each method averaged across three test objects, with shaded regions indicating standard deviation. The plot shows that our method achieves significantly lower standard deviation in Chamfer distance, indicating more reliable and consistent behavior)}
  \label{fig:plots}
  \vspace{-.4cm}
\end{figure*}

\noindent\textbf{Result 1: Our method enables effective tactile-driven exploration for object localization and identification.}
Despite the challenging setting where three objects are present simultaneously 
and the agent must locate a specific target while avoiding false 
identifications of distractors, our method consistently outperforms both \bconly and \heuristics baselines in terms of reconstruction accuracy, consistency, and object identification success. As shown in Table~\ref{tab:main_results}, we achieve higher success rates and lower Chamfer-$L_2$ distances across all objects. Fig.~\ref{fig:exp} illustrates representative test runs, while Fig.~\ref{fig:plots}(a) shows a steady decrease in reconstruction error over training episodes. 

The \heuristics baseline, which follows predefined exploration patterns without learning from contact feedback, exhibits high variance in contact coverage and reconstruction quality. Although it occasionally succeeds through incidental dense contact, it lacks consistency across trials and object geometries. In contrast, our method achieves significantly lower variance in Chamfer distance, indicating more reliable and repeatable behavior (Fig.~\ref{fig:plots}(b)).

Compared to the \bconly policy, which is constrained by offline expert demonstrations, our policy improves performance through on-policy adaptation guided by structured reward signals. BC initialization gives RL a structured prior over safe contact strategies, avoiding costly exploration from scratch on real hardware. This behaviour also explains why our policy performs better than the \rlnobc baseline, as it allows the policy to focus on refining exploration and reconstruction behaviours rather than discovering viable contact strategies from scratch, which is time-consuming in real-world setups.

Qualitatively, the learned policy produces contact trajectories that target geometrically informative and previously unexplored regions, enabling the shape completion model to generate more accurate reconstructions (Fig.~\ref{fig:exp}). This leads to earlier and more confident episode termination based on reconstruction quality. Despite allowing independent control of all six finger joints, the policy exhibits coordinated grasp-like motions interleaved with sliding and grazing behaviors (see supplementary video). Such coordination emerges naturally from BC initialization and reward shaping and yields richer tactile information per action, particularly on curved or convex surfaces.

\begin{table}[htp!]
\centering
\caption{\textbf{Ablation study on the effect of different reward components.} All three rewards are essential; removing any leads to reduced exploration, contact quality, or reconstruction accuracy.}
    \begin{tabular}{l c c}
    \toprule
    Reward Components & Success $\uparrow$ & Chamfer-$L_2$ (m) $\downarrow$ \\
    \midrule
    $r_{\text{recon}} + r_{\text{contact}}$ & 0.33 & 0.045$\pm$0.025 \\
    $r_{\text{recon}} + r_{\text{explore}}$ & 0.41 & 0.054$\pm$0.032 \\
    $r_{\text{explore}} + r_{\text{contact}}$ & 0.50 & 0.060$\pm$0.041 \\
    $r_{\text{recon}} + r_{\text{contact}} + r_{\text{explore}}$ & \textbf{0.77} & \textbf{0.015$\pm$0.010} \\
    \bottomrule
    \end{tabular}
\label{tab:rewards}
\vspace{-.3cm}
\end{table}

\noindent\textbf{Result 2: All three rewards are necessary for effective exploration and object reconstruction.} Our results, Table~\ref{tab:rewards}, indicate that all three rewards are necessary to effectively guide the agent through the stages of workspace exploration, object interaction, and shape reconstruction. Removing the $r_{\text{explore}}$ causes the agent to remain near its initial configuration, leading to failures unless an object is encountered early. Excluding $r_{\text{contact}}$ results in poor surface coverage despite successful object discovery, while omitting $r_{\text{recon}}$ yields partial reconstructions with higher geometric error. Only the full reward formulation consistently achieves high success rates and low reconstruction error, confirming the necessity of jointly balancing exploration, contact quality, and reconstruction accuracy.

\begin{table}[htp!]
\centering
\caption{\textbf{Ablation study on the role of shape completion.} Shape completion improves object identification success and reconstruction accuracy, reducing ambiguity from sparse tactile inputs.}
    \begin{tabular}{l c c}
    \toprule
    Method & Success $\uparrow$ & Chamfer-$L_2$ (m) $\downarrow$ \\
    \midrule
    No shape completion & 0.34 & 0.038$\pm$0.050 \\
    With shape completion & \textbf{0.77} & \textbf{0.015$\pm$0.010} \\
    \bottomrule
    \end{tabular}
\label{tab:shape}
\vspace{-.3cm}
\end{table}

\noindent\textbf{Result 3: Shape completion model is essential for robust object identification.}
As shown in Table~\ref{tab:shape}, removing the shape completion model significantly reduces object identification success, even when the Chamfer distance appears low. Sparse tactile point clouds can be geometrically ambiguous and may incidentally match multiple objects, leading to premature episode termination without correct identification. The learned shape completion model provides a strong prior that resolves this ambiguity by producing dense, object-specific reconstructions, resulting in substantially improved success rates and reconstruction accuracy.

\section{Limitations}
While our framework demonstrates effective tactile exploration and object localization in real-world settings, several limitations warrant discussion. Our method assumes operation from a known library of three object models. While our experiments demonstrate the method's core mechanisms, broader validation on larger object sets (e.g., YCB benchmark objects) and more complex geometries would strengthen generalization claims. Real-world data collection remains costly, limiting our experimental scope. Furthermore, our experiments focus on static objects, as estimating object displacement or pose changes without vision is challenging. Extending the framework to dynamic environments can be addressed by integrating visuo-tactile sensing to infer object displacement.

\section{Conclusion}
In this work, we presented \ourmethod, a tactile-based framework for autonomous exploration and object identification in environments without visual perception. Our policy trained in the real-world enables a multi-fingered robot to \textit{explore} an unknown workspace, \textit{localize} objects through contact, and \textit{reconstruct} their 3D geometry without any visual input. Our results demonstrate that our method outperforms baselines in tactile exploration and object identification through reconstruction in real-world settings. Quantitatively, we observed an overall success rate of 77\% with an average Chamfer-$L_2$ loss of 0.015 m. Qualitatively, we observe that the agent learns to explore the workspace effectively to perform purposeful contact interactions that lead to accurate reconstructions. These results highlight the potential of tactile sensing not only as a reactive fallback for vision but as a primary modality for perception in unstructured, occluded, or confined environments. Future work will extend our system to dynamic scenes and improved generalization.

\addtolength{\textheight}{-12cm}   




\begingroup
\small
\bibliographystyle{IEEEtran}{}
\bibliography{references}

\end{document}